\documentclass{article}
\usepackage{acra}

\usepackage{amsmath}

\usepackage{hyperref}
\usepackage{gensymb}

\usepackage{booktabs}
\usepackage{makecell}
\usepackage[pdftex]{graphicx}

\title{Low-cost Real-world Implementation of the Swing-up Pendulum for Deep Reinforcement Learning Experiments}

\author{Peter B{\"o}hm\thanks{Corresponding Author. The research for this paper received funding support from the Queensland Government through Trusted Autonomous Systems (TAS), a Defence Cooperative Research Centre funded through the Commonwealth Next Generation Technologies Fund and the Queensland Government}, Pauline Pounds, and Archie C. Chapman  \\ The University of Queensland, Brisbane, Australia \\ 
	p.bohm@uq.edu.au, pauline.pounds@uq.edu.au, archie.chapman@uq.edu.au}

\begin{document}

\maketitle

\begin{abstract}
Deep reinforcement learning (DRL) has had success in virtual and simulated domains, but due to key differences between simulated and real-world environments, DRL-trained policies have had limited success in real-world applications.
To assist researchers to bridge the \textit{sim-to-real gap}, in this paper, we describe a low-cost physical inverted pendulum apparatus and software environment for exploring sim-to-real DRL methods. 
In particular, the design of our apparatus enables detailed examination of the delays that arise in physical systems when sensing, communicating, learning, inferring and actuating.
Moreover, we wish to improve access to educational systems, so our apparatus uses readily available materials and parts to reduce cost and logistical barriers.  
Our design shows how commercial, off-the-shelf electronics and electromechanical and sensor systems, combined with common metal extrusions, dowel and 3D printed couplings provide a pathway for affordable physical DRL apparatus.
The physical apparatus is complemented with a simulated environment implemented using a high-fidelity physics engine and OpenAI Gym interface.
Our experiments show that the real-world environment is not solvable using plain DRL algorithms (DQN, TD3) and requires use of techniques for engineering a ``learnable'' state representation. 
In contrast, the simulated environment is solvable using the same plain algorithms, thereby illustrating the sim-to-real gap.

\end{abstract}

\begin{figure}
	\centering
	\includegraphics[width=0.4\textwidth]{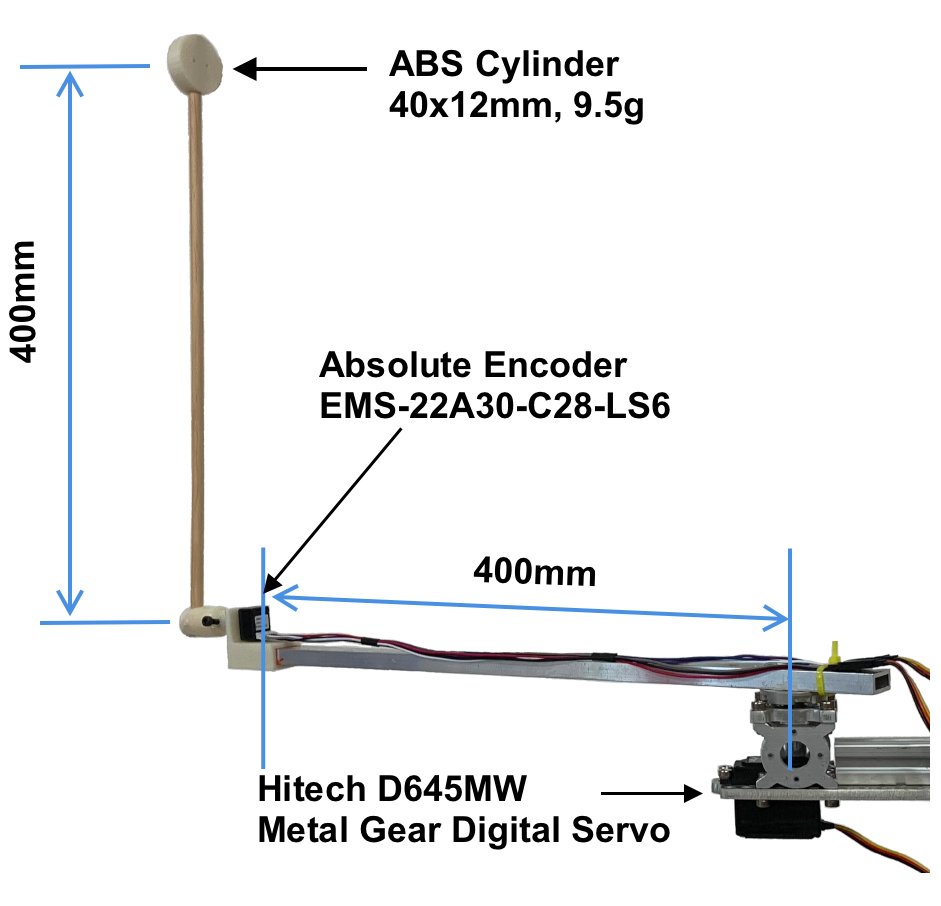}
	\caption{Pendulum apparatus technical details.} 
\label{fig:rig-dims}
\vspace{-2mm}
\end{figure}

\section{Introduction}
Inverted pendulums have long been used in control theory experiments, and have been popular in engineering education since the 1950s. 
Although they are simple, they are used to demonstrate a multitude of concepts such as linear control, non-linear control and feedback stabilization as well as hybrid systems and control of chaotic systems \cite{aastrom2000swinging,lundberg2010history}.
These principles apply in a wide variety of important and valuable domains, including robotic automation, aeronautics, and manufacturing and materials plant process control. 

Given their importance and prominence, inverted pendulums have also become one of the classic experiments in \textit{deep reinforcement learning} (DRL) training.
The Pendulum Swing-up and CartPole incarnations of this problem appear in most DRL test suites\cite{brockman2016openai,coumans2019,tassa2018deepmind}.
However, these idealized, simulated implementations are easy to solve because their accurate virtual sensor readings and instantaneous observations make for a predictable and highly-observable environment that reinforcement learning can readily and quickly learn from.


Clearly, these simulated environments are not a true reflection of physical systems in the real world; the \textit{sim-to-real gap} remains a considerable barrier to deploying DRL systems in meaningful applications. 
A major part of this gap is the large difference between simulated experimental RL setups and the actual behavior of real-world robotics systems \cite{dulac2019challenges,ibarz2021train}. 
Some important aspects of this gap include: 
the presence of delays in sensing an computational components of real-world DRL environments; 
inaccurate and otherwise limited sensor measurements; imprecise actuation of controllable elements, and; 
changes in physical machines due to mechanical fatigue, wear and tear.
We have observed that a key feature of real-world environments is that they do not stop while training and inference are conducted or wait for communication to occur. Rather their state evolves continuously, therefore DRL methods must be robust to sources of delay and other imperfections in state observations and actuation.
Moreover, accurately assessing the effects of these real-world features and the delays they entail requires building physical hardware to test DRL methods on.

In this paper, we describe a low-cost physical inverted pendulum apparatus and software environment for exploring DRL training on hardware and developing sim-to-real DRL methods. 
In particular, the design of our hardware apparatus enables detailed examination of the delays which exist in sensors, actuators, and communication channels, as well as control signals' delays due to time spent computing inference and learning.
Our approach is to use off-the-shelf, commodity grade components, in which imperfections in the learning environment quickly rise to prominence; 
to employ existing open-source software elements enabling straightforward extension of existing DRL routines to physical experiments; and to expose as many of the sensing and computational routines to direct measurement and logging, to enable accurate performance analysis and refinement.

The simulated environment is implemented using the MuJoCo physics engine \cite{todorov2012mujoco} and provides a digital clone of the physical apparatus that can be used for pre-training, sim-to-real transfer or to study differences between simulated and real-world environments.

While developing this hardware and software, we are driven to increase the accessibility of educational systems by curtailing costs.  
By using readily available materials and parts, both cost and logistical barriers to assembling these apparatuses can be reduced.  
Commercial off-the-shelf electronics and electromechanical and sensor systems, combined with common metal extrusions, dowel and 3D printed adaptors provide a pathway for very affordable implementations.  However, low cost assemblies typically come at the cost of irregular construction and variable performance; generally, educational hardware is constructed to a high standard of finish to simplify behaviour and reduce the burden on the student.  
Paradoxically, in the case of DRL systems, the complexity and imperfection of apparatus offers an advantage: the systems encapsulate sufficiently rich behaviour to demonstrate the advantages of reinforcement learning in systems difficult to regulate from first principle models, providing fertile ground for teaching and discussion beyond the specific case of a swing-up pendulum.

The remainder of the paper is as follows.
Section~\ref{sec:impl-hdd} describes the hardware components of our swing-up pendulum apparatus.
Section~\ref{sec:impl-soft-controller} describes the control loop used to implement DRL on the swing-up apparatus, while Section~\ref{sec:impl-gym} explains Gym Environment, which is the software intermediary that sits between the hardware micro-controller and the DRL routines. 
In Section~\ref{sec:r-dqn}, we describe a dimension reduction technique that acts as an interface between the raw sensor data streams and the DRL methods, which we have found is an essential component of the apparatus. 
Finally, Section~\ref{sec:educational} discusses educational uses for the apparatus, including how our apparatus can be used as a teaching aid or a student project, and provides some expected student outcomes.

\section{Implementation - Hardware}
\label{sec:impl-hdd}
Our aim is to create a minimal model apparatus that allows us to study the real-world behaviour of electronic and electromechanical systems, such as delays and noisy sensor readings unmediated by complex design or extraneous components/moving parts.
Given our broader goal of developing methods and techniques that allow us to apply DRL to real-world problems, all of our components are low cost and consumer-grade.

The pendulum apparatus is shown in Figure~\ref{fig:rig-dims}. 
It consists of an aluminium rectangular tube attached to a servo with a rotary encoder mounted at the other end. 
The pendulum consists of a pine dowel with a weight and is coupled directly to the encoder.
To provide more space for the moving pendulum, the servo is mounted on an aluminium support clamped to the table.

The system is driven by a micro-controller that reads the position from the encoder and commands the servo through a servo shield. 
Communication with the host computer uses an MQTT\footnote{https://mqtt.org/} protocol based message broker over WiFi network. 
Both the MCU and the servo are powered by a bench power supply using two separate channels with different voltage and maximum current.

\subsection{Arm}
The arm is made of a rectangular aluminium tube 20x10mm with 2mm thick sides. 
The length from the mounting point to the servo's shaft to the end of the arm is 400mm.

Initial random exploration of the DRL training may cause the arm to abruptly change direction at a high frequency, 
thus the material needs to be strong enough to withstand the resulting forces exerted on the arm.
Although there are DRL techniques designed for safe exploration of the action space, this more resilient implementation allows for study of larger variety of algorithms and exploration techniques.

We experimented with several other materials including 8mm diameter aluminium tube and 3D printed arm, but none was strong enough to withstand the load.

\subsection{Pendulum}
The pendulum is a 400mm long pine dowel. 
The weight at its end is a 3D printed cylinder, 40mm in diameter, 12mm high, with mass of 9.5g.
The pine dowel is lightweight and slightly flexible. 

We also experimented with aluminium tube. However, aluminium lacks flexibility and the tube eventually failed.
The compliance of the dowel prevents accumulation of fatigue.

\subsection{Encoder}
The pendulum is coupled to a rotary encoder. There are two main groups of rotary encoders: (i) absolute, and (ii) incremental. 
Absolute encoders generate a unique code for each shaft position, which represents the absolute position of the encoder. 
Incremental encoders generate an output signal each time the shaft rotates an angle (given by the encoder's resolution; Some also generate an additional signal on full rotation). 
The choice of encoder influences some of the calculations, but in general, either of the type could be used. 

We used an absolute encoder with a resolution of 1024 positions (a low-cost Bourns encoder EMS-22A30-C28-LS6). 
The rotary encoder measures the pendulum's angle $\theta$ with zero at the bottom and 512 at the top.

\begin{figure*}[h!]
	\minipage{0.32\textwidth}
	\centering
	\includegraphics[width=0.9\textwidth]{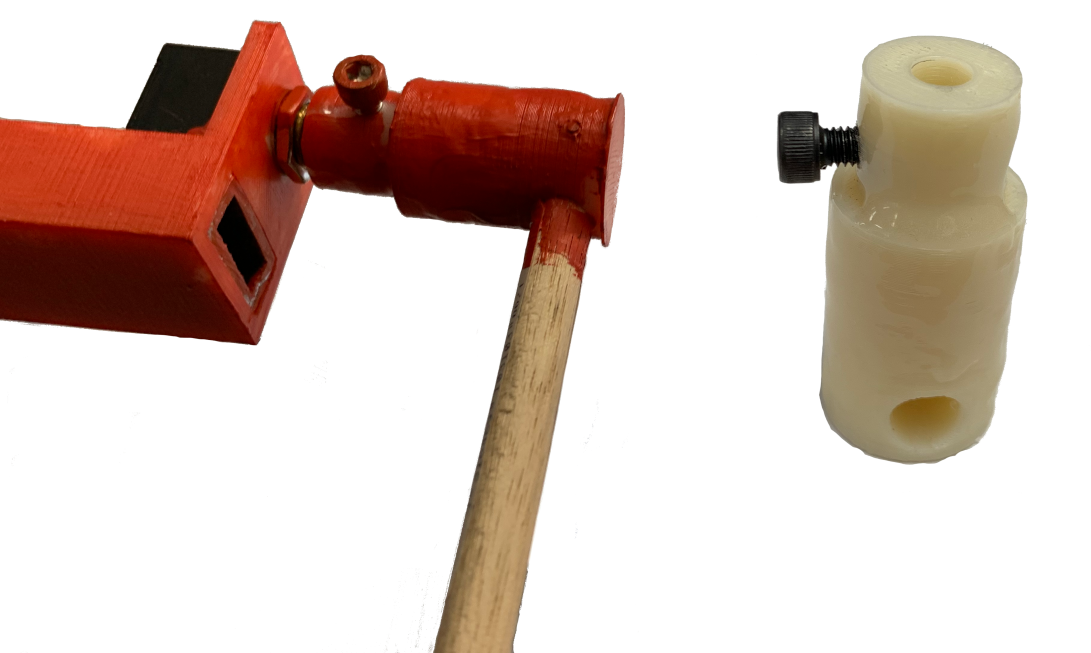}\\
	\caption{3D printed pendulum coupler.}
	\label{fig:coupler}
	\endminipage\hfill
	\minipage{0.32\textwidth}
	\centering
	\includegraphics[width=0.5\textwidth]{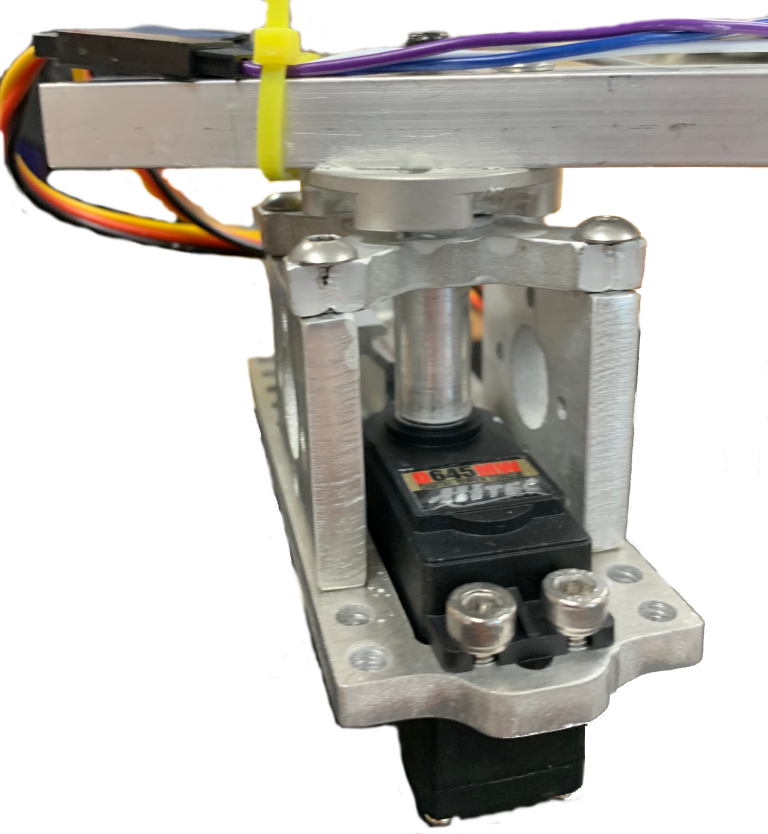}
	\caption{Servo block with a long shaft hub.}
	\label{fig:servo-block}
	\endminipage\hfill
	\minipage{0.32\textwidth}
	\centering
	\includegraphics[width=0.6\textwidth]{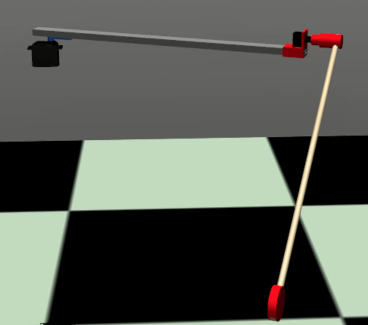}
	\caption{Simulated environment using MuJoCo physics engine.}
	\label{fig:pendulum-r-sim}
	\endminipage\hfill
\end{figure*}

\subsection{Servo}
To move the arm, we use a metal gear digital servo (Hitec D645MW). This is a standard hobby grade servo with stall torque of 8.3kg/cm at 4.8V (11.3kg/cm at 6V) and speed of 0.28sec/60$^{\circ}$ at 4.8V (0.20sec/60$^{\circ}$ at 6V).

Both speed and torque are important aspects of servo selection. 
The speed determines the limits of the angular velocity and influences the control step size. To make the pendulum controllable, the angular velocity cannot be too high or too low. It should also allow for control input of different speeds (by varying the angular increment size).   

The holding torque also contributes to the stability of the system. The torque needs to be high enough to stop the arm at the required position, however, a very high torque would stop the pendulum too abruptly disrupting its movement. We tested this with a high-torque servo Hitec HSB-9380TH. The abrupt starts and stops induced oscillation, leading to uncontrollable behavior and incipient system failure.

\subsection{Coupling}
There are two points that require joining: (i) the pendulum is coupled to the encoder, and (ii) the arm is coupled to the servo. 
For the pendulum, we used a 3D printed coupler, as shown in Figure~\ref{fig:coupler}. 
We used ABS filament. 

The earlier versions of this design developed cracks around the locking screw, eventually causing them to break. 
This breaking was reduced by epoxy coating and using a larger, M4 sized screw. 

Joining the arm to the servo shaft needs to support the weight of the arm with the encoder and pendulum. The arm acts as a lever adding extra stress to the mounting point.

We had the best results with a servo block consisting of a bearing-supported long shaft servo hub, which alleviates the radial load exerted on the servo shaft, as shown in Figure~\ref{fig:servo-block}.
A simpler alternative is anodised aluminium alloy horn with clamping style head for secure mounting. 
Still, the horns are damaged through wear and tear and need to be changed regularly. 
One of the indicators of fatigue is the right-angle between the shaft and the arm reducing by more than 1 or 2 degrees.

\subsection{Power Supply}
Both the MCU and the servo shield are powered using a bench power supply. Each uses a separate channel: the MCU requires 5V restricted to 1A current and the servo shield is powered by 6V and 3A. Servos are rated with different torque/speed at different voltages which opens up opportunities for further experimentation.

\subsection{Microcontroller (MCU)}
To control the system, we used SparkFun's Thing Plus ESP32S2. 
The servo is controlled through a servo shield (Adafruit PCA9685).
This controller was chosen because it is WiFi capable allowing for asynchronous and non-blocking communication via a message broker based on MQTT protocol.

We also experimented with Arduino Uno and Teensy 4.0. However, we ran into several issues during some of the experiments: (i) the Arduino runs at 16MHz compared to 600MHz for Teensy 4.0 (and 240MHz for ESP32), and this was causing delays when using faster action updates and/or sensor streaming frequencies; 
(ii) Teensy 4.0 doesn't support Ethernet and required serial UART connection. 

\begin{table}[t]
	\caption{Swing-up pendulum hardware components.}
	\label{tab:my_label}
	\begin{center}
		\begin{tabular}{p{0.12\textwidth}p{0.31\textwidth}}
			\toprule
			MCU & 
			Thing Plus ESP32S2 \\ 
			Servo Shield &
			PWM/Servo Driver with I2C interface PCA9685 \\ 
			Encoder & 
			EMS-22A30-C28-LS6 \\
			Servo	&	Hitec D645MW
			\\ \bottomrule
		\end{tabular}  
	\end{center}
\end{table}

\section{Implementation: Control Loop}
\label{sec:impl-soft-controller}
The micro-controller code was developed using C++ and PlatformIO framework. 
It has the following functions: (i) communication with the training program (receiving actions and sending observations), (ii) providing safety overrides, (iii) executing the actions, and (iv) processing the sensor (encoder) readings and calculation of angular velocity and acceleration.

\subsection{Communication}
Communication with the DRL training running on the host machine has two parts: (i) receiving the actions, and (ii) sending out observations.
Time intervals between observation streaming and executing received actions are set via hyper-parameters.
The encoder is read and calculations updated each time the observations sending is triggered.

When the acting is triggered, the last received action is passed through the safety overrides and actuated.
In discrete action mode, the last received action will continue to be executed at the subsequent runs until either the next action is received or the arm reaches the min/max angle.
In continuous action mode, the action represents the servo position and the servo will remain in that position until the next action is received.

Both observation streaming and acting operate independently at their own frequencies.
This is in contrast to usual ``synchronous'' RL setup, where after executing the action, observations reflecting the state change caused by this action are returned.
This is important because this type of streaming communication is typically used in many real-world robotic applications (e.g. autopilot software such as ArduPilot or PixHawk use MAVLink to stream data).



\subsection{Safety Overrides}
There are two safety overrides implemented: (i) when the arm reaches maximum/minimum angle, and (ii) when the pendulum spins with angular velocity over 2 rotations per second.
In both cases, the action will be set to stop. This check happens directly before actuation and it is not possible for the commands to override the safety.

While the pendulum experiment is relatively safe, the second override was implemented after several incidents where the pendulum's velocity caused it to break off (in one case breaking the encoder mount and in another breaking the coupler). In other cases it lead to servo failures.
The number of rotations per second override is adjustable through hyper-parameters and can be disabled if required. 


\subsection{Calculations}
The encoder and servo positions are the only sensory observations. 
We include calculated angular velocity and acceleration observations for the pendulum and angular velocity of the servo arm to complement the observations.

Angular velocity is derived as a rate of change in position over some time $\Delta t$.
When using a rotary encoder\footnote{We are using an absolute encoder, however, both absolute and incremental encoders have their own challenges when deriving angular velocity.} to determine the position changes $\Delta \omega$, there are several problems:
\begin{itemize}
	\item When the subsequent measurements read 0 and 256, the encoder could have moved 256 positions, or 1024 - 256 positions; or $n \times 1024 + 256$ positions; or it could have moved from 0 to 100, then to -100 and only then to 256.
	\item Positions 0 and 1023 are next to each other causing discontinuity in readings.
\end{itemize}

The first issue can be mitigated by using short enough time interval, and with some assumptions on maximum velocity, by always taking the shorter interval. 
In more detail, there are two approaches: (i) time based that measures the time between fixed position movements, and (ii) position based that measures position change during a fixed time. 
Our absolute encoder doesn't generate signals on state changes and needs to be polled for current position, which rules out the first option. 
If the time interval is too short, however, many of the subsequent readings will be the same, leading to zero velocity.

In our calculations, we assume the shorter interval is correct. The final reading is further smoothed out by averaging the last three velocities. 

The discontinuity is identified by large values of $\lvert \Delta \omega \rvert$ indicating crossing of the 0/1023 position in which case the value is calculated as $\Delta \omega = 1024 - \lvert \Delta \omega \rvert$.


\section{Implementation: Gym Environment}
\label{sec:impl-gym}
We provide two environments: (i) PendulumR that controls the real-world apparatus, and (ii) PendulumR-Sim that controls the simulator. Both provide the same interface; the simulated version directly controls the simulator that runs on the same physical machine, whereas the real-world version
runs on the host computer, and acts as an intermediary between the DRL agent/training and the micro-controller controlling the hardware. 

Different training suites provide slightly different APIs \cite{brockman2016openai,tassa2018deepmind,juliani2018unity}, however, the main features are the same. 
The base functionality includes accepting an action and returning the new state observations along with the reward.
By using this standard API, the same DRL training can work across different environments.
Our implementation follows the OpenAI Gym\cite{brockman2016openai} format by implementing their \textit{Environment} interface.

The actual implementation is split into two components: (i) the DRL user (Gym) interface, and (ii) the controller/driver. 
The first interacts with the DRL training, by accepting the actions, parsing the raw observations received from the controller to a floating point vector, calculating the reward and keeping count of the episodes steps.
The second maintains connection to the MQTT message broker, and uses it to send out the action messages translated to commands and receive the raw observations.

\subsection{PendulumR Environment}
The environment provides the standard API used by the DRL training. 


\subsubsection{Observation Space} 
The observation space consists of (i) two sensory observations (the pendulum's encoder readings and servo position), (ii) calculated values for pendulum's angular velocity and acceleration, servo arm's angular velocity, (iii) time since the last action, and (iv) observation age (measured from the time they were received until they were used for inference).
The settings file allows to include/exclude the individual calculated values.

\subsubsection{Action Space} 
We provide both continuous and discrete action space implementattheions.
The action is a single value that actuates the servo.

In discrete space, there are five possible actions: (i) stop, (ii) move left by 1 position, (iii) move left by 2 positions, (iv) move right by one position, and (v) move right by 2 positions\footnote{The actual number of position can be adjusted based on the desired speed.}.

In continuous space the action is a value from interval [-1, 1], representing the leftmost and rightmost positions of the servo. To prevent too large movements, this value $u_t$ is passed through exponential filter resulting in the actuation command $\bar{a}_t = \bar{a}_{t-1} * c + u_t * (1 - c)$. The filter coefficient $c$ is set in the configuration file (we achieved best results with $c = 0.85$).

\subsubsection{Reset}
This method is called at start of each episode and in simulated environments it usually resets the state into the initial (often random) position.
Our implementation stops the servo for 2s, which leaves the pendulum at a random position moving with a random inertia (depending on its movement at the end of the last episode).

\subsubsection{Step}
This method implements the actual exploration step by actuating the received action and returning resulting state observations along with the calculated reward.
Here, it means the actions are passed to the controller that sends them out to the MCU, the executing thread sleeps for the step time delay, gets current observations from the controller and calculates the reward.
Unlike in simulated environments, where the next observations are calculated and immediately returned, with physical systems the step method needs to actually wait for the action to take effect resulting in new observations.

Importantly, because the actuation happens not only out of process but out of system (i.e. on the physical hardware), this process is very different from a simple method call in simulated environments.
The communication channels are noisy and have latency that could significantly delay the sensor readings containing the next state. 
More over the communication step cannot be blocking on the physical system itself because it would leave it uncontrollable for the duration of the block and it could also lead to buffering of the received actions that could leave the system unresponsive for a relatively long time.

To avoid these problems, our step method functions asynchronously. The actions are sent out to the physical environment without any blocking wait for the response. The function then sleeps for the step time (defined in the hyper-parameters) and reads the current observations from the controller.


\subsubsection{Reward Calculation}
The reward function used is: 
\[R = -\theta^2 - 0.5 \times \omega^2\] 
where $\theta$ is an angle measured from the upright position in radians and $\omega$ is revolutions per second.
This is similar to the function used by the OpenAI Gym's Pendulum environment but with different weight of the $\omega^2$. We also omit the control input because unlike in the OpenAI's environment, the control input is servo position (continuous action space) or direction of movement (discrete) and not motor torque that directly translates into system's energy.

\subsubsection{Episode Length}
Episode length is configurable, however, it needs to be long enough for the exploration process to reach the less accessible steps that (during the early stages of the training) require multiple swings there and back before the pendulum has enough energy to swing-up.
We used episode length of 500 steps, which takes about 28s to complete.

\subsection{Controller/Driver}
The driver implements communication with the MCU using a MQTT protocol based message broker over WiFi network. MQTT is a lightweight publish/subscribe messaging transport with minimal code footprint and network bandwidth commonly used in the Internet of Things (IoT) applications.

There are two message queues, one for observations and one for actions. The MCU subscribes to the actions queue and publishes to the observation queue. The controller/driver subscribes to the observations queue and publishes to the actions queue. The communication is fully asynchronous with both sides publishing messages independently and without blocking.

This architecture is also easily scalable to multiple physical devices\footnote{Shown in the accompanying video.} with each device and exploration agent pair using separate message queues. The collected transitions are stored in the same experience replay buffer and used by a shared DRL training process.

\subsubsection{Sending Actions} 
Action commands include both actuation commands for controlling the servo and configuration commands to update hyper-parameters/settings without requiring re-flashing of the MCU.
In discrete mode, the actuation command has a form of $ma$ where $a$ is a number from interval [0, 4] representing one of the possible actions.
In continuous mode, it has a form of $ba$ where $a$ is a real number from interval [-1, 1] representing normalized position of the servo.

The configuration commands include adjusting encoder offset, observation streaming delay, acting delay, switching between discrete and continuous mode, safety override value, and starting/stopping of the observation streaming.


\subsubsection{Processing Streaming Observations} 
The entire process of checking and parsing the received observations runs on a detached background thread that implements MQTT message receiver. 

The MCU sends the observations as a comma separated list of values containing reference time in milliseconds, encoder position, calculated angular velocity and acceleration of the pendulum and angular velocity of the arm.
The controller parses these messages and the raw observations are provided to the PendulumR environment and used in the \textit{step()} function.

\subsection{Simulated Environment}
The simulated environment\footnote{Available at \url{https://github.com/real-world-drl/pendulum_r_sim}.} implements the same OpenAI Gym interface and the actual simulation is implemented in MuJoCo physics engine.
All dimensions closely follow the physical apparatus and the digital clone is composed of the same 3D meshes used in the original design and 3D printing (see Figure \ref{fig:pendulum-r-sim}).

In the baseline implementation without simulating delays, each steps takes 12 frame skips of $5\mathrm{ms}$ each, resulting in $\Delta{t} = 60ms$ which is closely mimics the physical apparatus.
Nevertheless, the lack of delays and various measurement inaccuracies make this environment considerably easier to train on and eventually solve.

To bring this simulation closer to reality, we also implemented the environment with delays. 
In this version, each step takes 6 frame skips after which the state observations are collected and the environment moves for another 0, 1 or 2 frame skips (drawn from the uniform distribution).
This version is considerably more difficult to solve, with both DQN and TD3 not being able to solve it.

\section{R-DQN and R-TD3 Demonstrations}
\label{sec:r-dqn}
The purpose of the apparatus is to foster experimentation and as such there are various ways of approaching the solution. From training directly on the hardware, through training on the simulator and deploying sim-to-real techniques such as the dynamics randomization \cite{peng2018sim} to a hybrid solution combining observations collected from both the real world and simulator.

Here, we show one of the possible solutions.
We have developed a modified version of the DQN \cite{mnih2013playing} and TD3 \cite{fujimoto2018addressing} algorithms
in which the observations are pre-processed through a \textit{gated recurrent unit} (GRU) \cite{cho2014learning}.
The GRU network is placed outside of the DQN network, serving as a variable sequence encoder that reduces the dimensionality of the observations sequence from the start of the episode to a fixed length vector.
This vector is then combined with observations from the current step and the result is cached in the replay buffer and used during training.
Observations are encoded cumulatively, with each step further unrolling the hidden state of the GRU.
We denote this algorithms R-DQN and R-TD3 respectively.

To account for variable delays present in the real-world environments, the training was done using a non-blocking and asynchronous architecture described in \cite{bohm2022nonblocking}.

\begin{figure}
	\centering
	\includegraphics[width=0.5\textwidth]{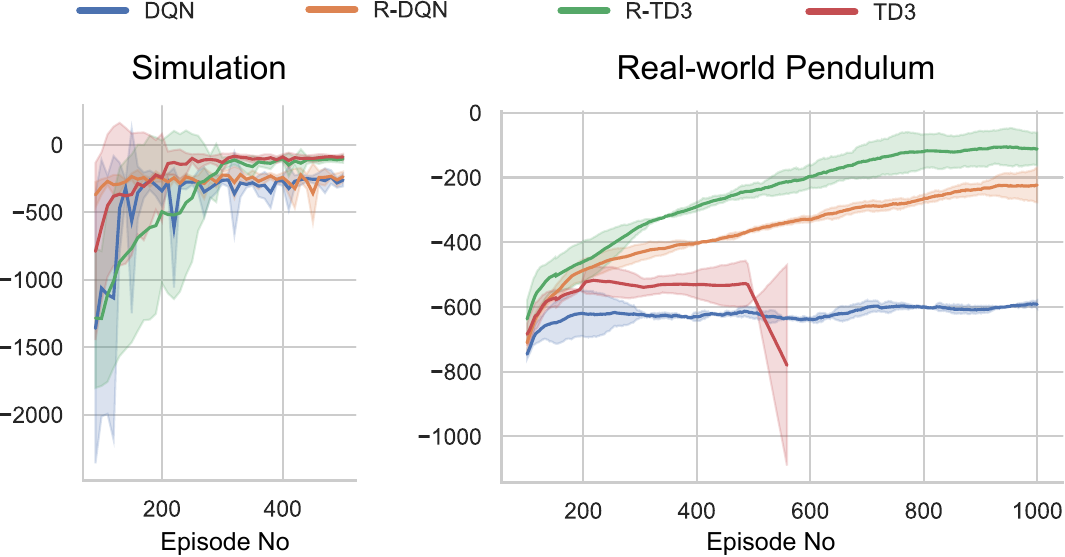}
	\caption{Learning curves for real-world and simulated environments. The simulated environment is easily solved by all algorithms, however, both plain versions of the DQN and TD3 fail in the real-world. TD3 training lead to servo failure in 2 out of 3 experiments.}
	\label{fig:pendulum-charts}
	\vspace{-2mm}
\end{figure}

\begin{figure*}[h!]
	\centering
	\includegraphics[width=0.8\textwidth]{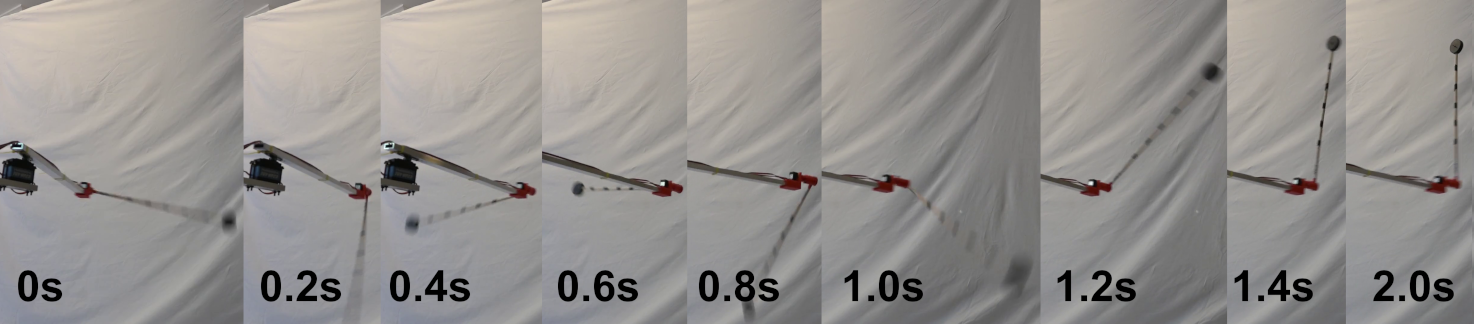}\\
	\hspace{2pt}\\
	\includegraphics[width=0.8\textwidth]{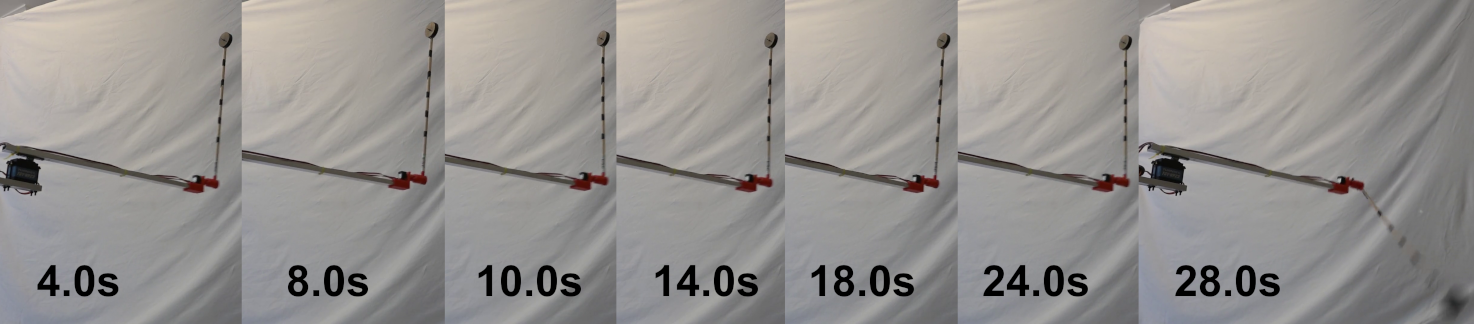}
	\caption{Pendulum swings into the upright position in 1.4s (equivalent to approximately 25 steps), and remains in the upright position until the end of the episode.}
	\label{fig:demo-shots}
\end{figure*}

The learning curves in Figure~\ref{fig:pendulum-charts} show the training results of both GRU preprocessed and plain algorithms in both environments. The real-world experiments consisted of three runs and simulated environment of ten runs each with a different, randomly generated seed. 
We were forced to stop the real-world TD3 experiments after about 500 episodes because the aggressive exploration lead to servo failure in the first two runs.
The simulated experiments were run for 2000 episodes, however, the training plateaued after about 400 episodes and remained stable until the end with the pendulum upright throughout entire episodes.
By way of example, Figure~\ref{fig:demo-shots} shows the performance of our R-DQN algorithm when embedded in a novel asynchronous training architecture (a paper describing the details of this architecture is under review). 

These results confirm the gap between simulated RL setups and the actual behavior of real-world robotics systems, as well as the importance of real-world experiments. 
Further, this shows that off-the-shelf DRL algorithms may not be sufficient given the degree of noise and delay in the sensor data making the learning from it without any pre-processing very difficult, verging on impossible.

We have also successfully experimented with other servo brands: Turnigy TGY-WP23 and MKS X5 HBL550. Different resolution and holding torque in those servos produced visibly different policies. 
These policies were also not transferable among the servos, that is a policy trained on the Turnigy servo did not perform on the Hitec or MKS servo. These differences reveal subtle real-world complexities that further complicate sim-to-real transfer and simulation modelling.

\section{Educational uses}
\label{sec:educational}
For instructors and professors of intermediate-level courses, this apparatus aids in the teaching of the fundamentals of robotic systems. 
It enables demonstrations of various modeling and control techniques, including 
kinematics and dynamics, 
motion planning and control,
as well as the reinforcement learning target application described earlier.

Construction of the apparatus described above provides an ideal project task for advanced undergraduate and postgraduate students of mechatronics and robotic systems,
system integration and internet of things.
We argue that the system is simple enough to describe to be presented to undergraduate students, while the task of effectively integrating the system components presents a degree of complexity that is challenging enough for pedagogical development. 
Through constructing their own version of the swing-up pendulum apparatus, students will traverse the steps of design, implementation, testing, evaluation and presentation of mechatronic system of intermediate size and complexity.
Specific steps that students will have to go through include: (i) evaluating sensor/actuator placement and alternative component choices, (ii) developing sensing and control algorithms on a practical robotic system, (iii) and applying a systematic approach to the design process for robotic control and/or reinforcement learning systems.

Finally, this apparatus, and extensions to it, provides a useful real-world deep reinforcement learning training environment on which cutting-edge DRL techniques can be evaluated for their ability to span the sim-to-real gap.

\bibliography{bibliography}
\bibliographystyle{named}

\end{document}